\documentclass[preprint,12pt]{elsarticle}

\usepackage{amssymb}
\usepackage{amsmath}

\usepackage[T1]{fontenc}
\usepackage{graphicx}
\usepackage{colortbl}
\usepackage{color}
\usepackage[table]{xcolor}
\usepackage{microtype}
\usepackage{bibentry}
\usepackage{subfigure}
\usepackage{multirow}
\usepackage{rotating}
\usepackage{booktabs}
\usepackage{pdfpages}
\usepackage{caption}
\usepackage{lipsum}
\usepackage{subcaption}
\usepackage{algorithm}
\usepackage{algorithmic}
\usepackage{adjustbox} 
\usepackage{hyperref}

\journal{Neurocomputing}

\makeatletter
\def\ps@pprintTitle{%
  \let\@oddhead\@empty
  \let\@evenhead\@empty
  \def\@oddfoot{}%
  \let\@evenfoot\@oddfoot}
\makeatother

\begin{document}

\begin{frontmatter}



\title{Perceptual Influence: Improving the Perceptual Loss Design for Low-Dose CT Enhancement}

\author[ufape]{Gabriel A. Viana\corref{cor1}}
\ead{gabriel.almeidaviana@ufape.edu.br}

\author[ufape,uantwerp]{Luis F. Alves Pereira}
\ead{luis-filipe.pereira@ufape.edu.br}

\author[ufpe]{Tsang Ing Ren}
\ead{tir@cin.ufpe.br}

\author[ufpe]{George D. C. Cavalcanti}
\ead{gdcc@cin.ufpe.br}

\author[uantwerp]{Jan Sijbers}
\ead{jan.sijbers@uantwerpen.be}

\cortext[cor1]{Corresponding author}

\affiliation[ufape]{organization={Universidade Federal do Agreste de Pernambuco (UFAPE)}, 
            addressline={Av. Bom Pastor, s/n, Boa Vista}, 
            city={Garanhuns},
            postcode={55292-270}, 
            state={Pernambuco},
            country={Brazil}}

\affiliation[ufpe]{organization={Centro de Informática, Universidade Federal de Pernambuco (UFPE)},
            addressline={Av. Jornalista Aníbal Fernandes, s/n, Cidade Universitária},
            city={Recife},
            postcode={50740-560},
            state={Pernambuco},
            country={Brazil}}

\affiliation[uantwerp]{organization={University of Antwerp}, 
            city={Antwerp},
            state={Antwerp},
            country={Belgium}}

\begin{abstract}
Perceptual losses have emerged as powerful tools for training networks to enhance Low-Dose Computed Tomography (LDCT) images, offering an alternative to traditional pixel-wise losses such as Mean Squared Error, which often lead to over-smoothed reconstructions and loss of clinically relevant details in LDCT images. The perceptual losses operate in a latent feature space defined by a pretrained encoder and aim to preserve semantic content by comparing high-level features rather than raw pixel values. However, the design of perceptual losses involves critical yet underexplored decisions, including the feature representation level, the dataset used to pretrain the encoder, and the relative importance assigned to the perceptual component during optimization. In this work, we introduce the concept of \textit{perceptual influence} (a metric that quantifies the relative contribution of the perceptual loss term to the total loss) and propose a principled framework to assess the impact of the loss design choices on the model training performance. Through systematic experimentation, we show that the widely used configurations in the literature to set up a perceptual loss underperform compared to better-designed alternatives. Our findings show that better perceptual loss designs lead to significant improvements in noise reduction and structural fidelity of reconstructed CT images, without requiring any changes to the network architecture. We also provide objective guidelines, supported by statistical analysis, to inform the effective use of perceptual losses in LDCT denoising. Our source code is available at \url{https://github.com/vngabriel/perceptual-influence}.
\end{abstract}

\begin{graphicalabstract}
\includegraphics[width=\linewidth]{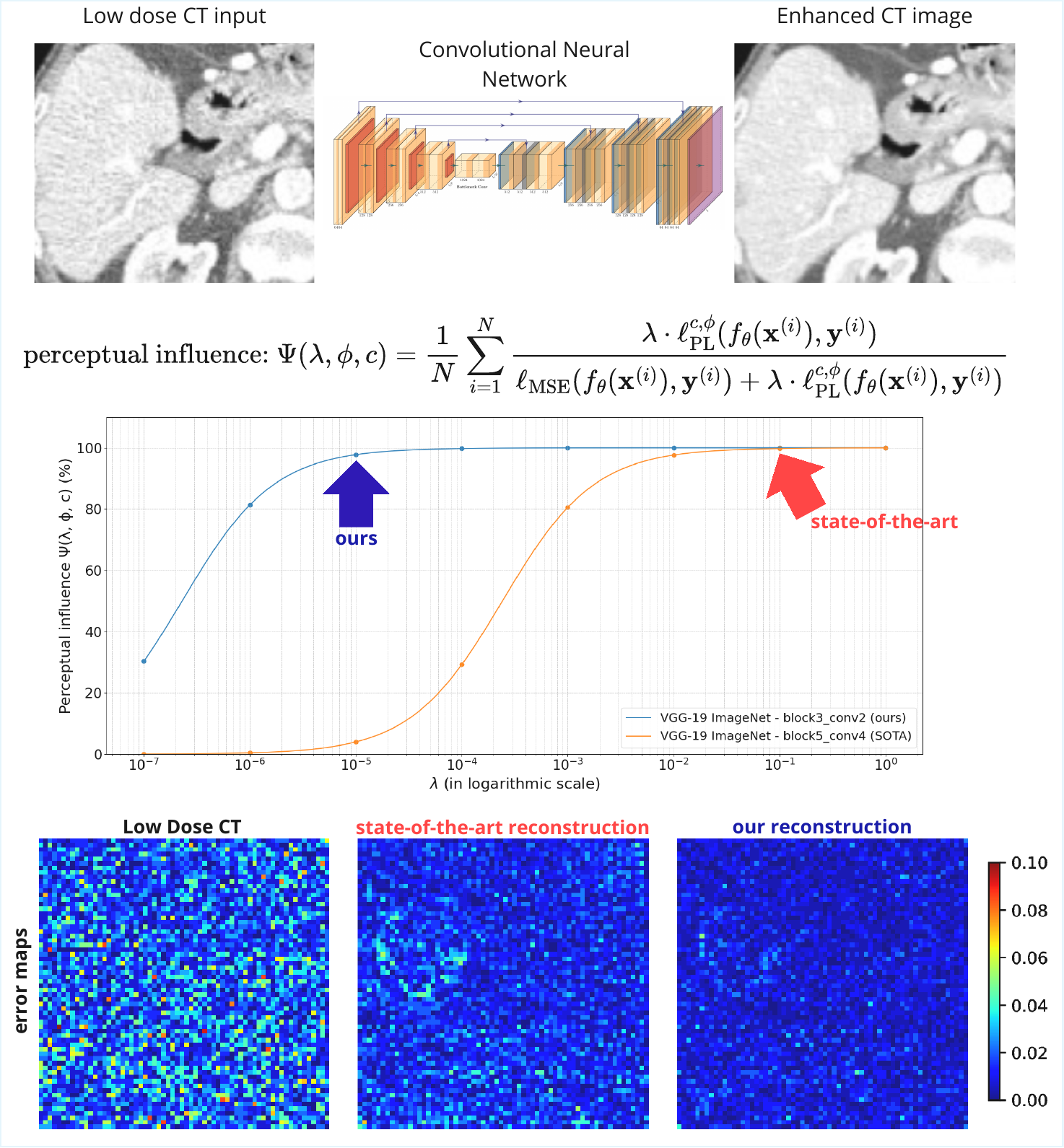}
\end{graphicalabstract}

\begin{highlights}
\item We introduce a new metric to assess the influence of perceptual loss in Low-Dose CT denoising
\item We investigate how the feature level and perceptual loss weight influence its effectiveness
\item We show that widely used configurations underperform compared to our optimized designs
\item Our methodological design improves noise reduction while preserving structural details in CT images
\item We offer guidelines for the effective use of perceptual loss in Low-Dose CT enhancement
\end{highlights}


\end{frontmatter}



\section{Introduction}

Denoising Low-Dose Computed Tomography (LDCT) images is a critical challenge in medical imaging, where the objective is to reduce radiation exposure while preserving the diagnostic quality of reconstructed scans. In recent years, Deep Learning (DL) has established itself as the state-of-the-art approach for this task \cite{7947200,8964295,ZHANG2024108112}, leveraging powerful neural architectures to learn mappings from noisy low-dose acquisitions to high-quality approximations of normal-dose images. These models are typically trained by minimizing a loss function that measures the discrepancy between the predicted and ground-truth images.

While pixel-wise losses such as Mean Squared Error (MSE) are widely used in LDCT denoising tasks due to their simplicity \cite{7947200}, they are known to produce overly smoothed outputs \cite{8964295,8340157}, suppressing high-frequency components and diminishing critical anatomical structures. To address this limitation, perceptual losses \cite{8340157,johnson2016perceptuallossesrealtimestyle,8099502} have been employed as complementary terms in the training objective. These losses operate in a latent feature space defined by a pre-trained encoder network, encouraging the model to preserve semantically meaningful content by aligning feature-level representations rather than raw pixel intensities.

Despite their effectiveness, the design of perceptual losses remains largely underexplored and is often performed in an ad hoc manner in the context of LDCT denoising. Key design parameters - such as the choice of encoder network, its pre-training context (e.g., natural vs. medical images), the feature extraction level, and the weighting factor that balances the perceptual and pixel-wise loss terms for training the network - are often adopted from precedent or empirical intuition rather than principled analysis. Our literature review \cite{8964295,8340157,li2025ctmambahybridconvolutionalstate,mahmoud2023variant,10535288,9023861,electronics13163184,HUANG2021104} shows that most studies focus on proposing new models and network architectures, leaving many questions unanswered regarding the design choices used to set up perceptual losses for enhancing LDCT images. This raises the concern that the current state-of-the-art relies on conventional rather than optimal configurations, potentially limiting performance of LDCT denoising models.

In this work, we address this gap by introducing a principled framework to investigate and improve the design of perceptual losses for LDCT enhancement. We propose the concept of \textit{perceptual influence} ($\Psi$), which is a metric that quantifies the relative contribution of the perceptual loss during training DL architectures to map LDCT images to their NDCT counterparts. By calibrating perceptual influence to remain consistent across configurations, we enable fair comparisons between different design choices, removing the confounding effects that arise when perceptual loss contributions are left uncontrolled.

To structure our investigation, we pose the following research questions:

\begin{enumerate}
    \item[\textbf{RQ1)}] To what extent does the \textit{perceptual influence} vary as a function of the selected feature representation level, across different pre-training contexts?
    
    \item[\textbf{RQ2)}] Which feature representation level yields the best enhancement results for different pre-training contexts?

    \item[\textbf{RQ3)}] Can objective guidelines be established for selecting appropriate configuration values to maximize the metrics?
\end{enumerate}

The remainder of this paper is organized as follows: In Section~\ref{sec:problem_state}, we introduce the role of perceptual loss in LDCT denoising and describe how its key configurations impact reconstruction quality. Section~\ref{sec:related_works} reviews related works and highlights the lack of systematic evaluation in current approaches. In Section~\ref{sec:methodology}, we present our proposed framework for improving LDCT denoising, centered on the concept of perceptual influence, which allows for principled comparison and calibration of loss configurations. Section~\ref{sec:experiments} details the experimental design, evaluating the impact of different encoder contexts and feature representation levels on denoising performance. In Section~\ref{sec:results}, we analyze the results through quantitative and qualitative metrics, assess the influence of each configuration, and discuss the broader implications of our findings. Finally, Section~\ref{sec:conclusion} concludes the paper by summarizing our contributions toward improving LDCT denoising, acknowledging limitations, and proposing directions for future work.

\section{Perceptual Loss in Low-Dose CT Denoising}
\label{sec:problem_state}

The denoising of LDCT images is commonly formulated as an optimization problem, where the goal is to recover high-quality images from noisy, low-dose counterparts. In this context, DL approaches generally model a mapping function $f_{\boldsymbol{\theta}}: \mathbb{R}^n \rightarrow \mathbb{R}^n$ \cite{ZHANG2024108112}, where $\mathbb{R}^n$ denotes the image space of dimension $n$. The objective is to identify the optimal set of parameters $\boldsymbol{\theta}$ that enables the mapping of a low-dose input image \( \mathbf{x} \in \mathbb{R}^n \) to a high-quality approximation \( f_{\boldsymbol{\theta}}(\mathbf{x}) = \hat{\mathbf{y}} \in \mathbb{R}^n \) of the corresponding normal-dose image \( \mathbf{y} \in \mathbb{R}^n \). This learned mapping facilitates the recovery of structural and textural details, which are typically degraded due to radiation dose reduction, thereby enhancing diagnostic image quality.

To train the model \( f_{\boldsymbol{\theta}} \) for this mapping, we aim to find the optimal parameter vector \( \boldsymbol{\theta} \) by minimizing the average loss function $\ell$ over a supervised dataset \( T = \{(\mathbf{x}^{1}, \mathbf{y}^{1}), \dots, (\mathbf{x}^{m}, \mathbf{y}^{m})\} \) of \( m \) input-output pairs, defined as:

\begin{equation}
    \boldsymbol{\theta} = \arg\min_{\boldsymbol{\theta}} \frac{1}{m} \sum_{i=1}^{m} \ell\left(f_{\boldsymbol{\theta}}(x^{i}), y^{i}\right).
    \label{eq:opt}
\end{equation}

A common choice for the loss function $\ell$ in such models is a pixel-wise loss, such as the Mean Squared Error (MSE), defined as:

\begin{equation}
    \ell_\text{MSE}(\mathbf{a},\mathbf{b}) = ||\mathbf{a}-\mathbf{b}||^2,
\end{equation}

\noindent where \( \mathbf{a} \) and \( \mathbf{b} \) are images in the same space \( \mathbb{R}^n\). While simple and intuitive, this formulation is known to produce excessively blurry outputs~\cite{you2018structurally,8964295}, often diminishing high-frequency details, which are essential for accurate clinical assessment.

To address this limitation, perceptual losses have been encouraged as alternatives \cite{johnson2016perceptuallossesrealtimestyle,8099502} to traditional pixel-wise losses. These losses operate in a latent feature space generated by a deep neural network encoder designed to extract meaningful representations from input images. Such an encoder can be expressed as a function \(\mathcal{E}: \mathbb{R}^n \rightarrow \mathbb{R}^d\), where \(d < n\) denotes the dimensionality of the latent feature space. This transformation enables the comparison of image content at a perceptual level rather than at individual pixel levels. A formulation of the perceptual loss is given by:

\begin{equation}
    \ell^{c,\phi}_{\text{PL}}(\mathbf{a}, \mathbf{b}) = \|\mathcal{E}^\phi_c(\mathbf{a}) - \mathcal{E}^\phi_c(\mathbf{b})\|^2,
    \label{eq:perceptual}
\end{equation}

\noindent where \(\mathcal{E}^\phi_c(\cdot)\) denotes the feature vector extracted from level \(\phi\) of the encoder \(\mathcal{E}\), which was pre-trained on a dataset/context \(c\). The index \(\phi\) controls the semantic level of the features \cite{8099502} - ranging from low-level textures to high-level structures - while the context \(c\) shapes the nature of the learned representations. This loss encourages the latent representation of $\mathbf{a}$ to align with the latent representation of $\mathbf{b}$, rather than merely minimizing differences at the pixel level.

Consequently, the total loss used in the optimization process often combines both pixel-wise and perceptual terms:

\begin{equation}
    \ell_{\text{total}}(\mathbf{a}, \mathbf{b}) = ||\mathbf{a}-\mathbf{b}||^2 + \lambda \|\mathcal{E}^\phi_c(\mathbf{a}) - \mathcal{E}^\phi_c(\mathbf{b})\|^2,
    \label{eq:total_loss}
\end{equation}

\noindent where \( \lambda \in \mathbb{R}^+ \) balances the contribution of the perceptual term.

\section{Related Works}
\label{sec:related_works}

In many state-of-the-art methods addressing LDCT denoising for learning $f_{\boldsymbol{\theta}}: \mathbb{R}^n \rightarrow \mathbb{R}^n$ using perceptual losses, the encoder ($\mathcal{E}$), is commonly implemented as a VGG-19 network \cite{simonyan2015deepconvolutionalnetworkslargescale,8340157,9023861,mahmoud2023variant,10535288}. Such VGG is typically pre-trained on a large-scale natural image dataset, most notably ImageNet \cite{5206848} ($c=ImageNet$), to learn rich and hierarchical visual features. With respect to the $\lambda$ value in Eq.~\eqref{eq:total_loss}, $0.1$ is a prevalent choice in several studies \cite{8964295,mahmoud2023variant,10535288,8340157}. Finally, the feature level $\phi$ is usually obtained from the \texttt{block5\_conv4} layer \cite{8340157}, which means the last layer from the VGG-19; i.e., the higher-level features. Table~\ref{tab:related_works} summarizes the choices made by various studies employing perceptual losses for directly mapping LDCT images to their NDCT counterpart, including the encoder $\mathcal{E}$, weighting factor $\lambda$, and feature extraction layers $\phi$.

\begin{sidewaystable}
    \centering
    \setlength{\tabcolsep}{7pt}
    \caption{Summary of Perceptual Loss Configurations in LDCT Denoising Literature}
    \label{tab:related_works}

    \resizebox{\textwidth}{!}{
    \begin{tabular}{llllll}
        \toprule
        \textbf{Reference} & $\mathcal{E}$ & $\lambda$ & $\phi$ & $c$ & \textbf{Notes on Justification/Choice} \\
        \midrule
        Yang et al. \cite{8340157} & VGG-19 & 0.1 & \texttt{block5\_conv4} & ImageNet & $\lambda$ chosed based on \textit{experimental experience}. \\
        \midrule
        Li et al. \cite{8964295} & Autoencoder & 0.1 & Last encoder layer & Medical & $\lambda$ chosed based on \cite{8340157} \\
        \midrule
        Marcos et al. \cite{10.3389/frsip.2021.812193} & VGG-16 & Adaptive & \texttt{block1\_conv2}, \texttt{block2\_conv2} & ImageNet & $\lambda$ adapted based on epoch loss. \\
        & & & \texttt{block3\_conv3}, \texttt{block4\_conv3} & & \\
        \midrule
        Trung et al. \cite{9023861} & VGG-19 & 0.1 & NS & ImageNet & $\lambda$ choice is not explained. \\
        \midrule
        Mahmoud et al. \cite{mahmoud2023variant} & VGG-19 & 0.1 & NS & ImageNet & $\lambda$ chosed based on \cite{8340157}.\\
        \midrule
        Zhang et al. \cite{electronics13163184} & Autoencoder & 0.1 & Last encoder layer & Medical & The choice of $\lambda$ lacks explanation.\\
        \midrule
        Saidulu et al. \cite{10535288} & VGG-19 & 0.1 & NS & ImageNet + & Fine-tuned for CT-specific features; \\
        & & & & Medical & Perform hyperparameter optimization;\\
        & & & & & $\lambda$ chosed based on \cite{8340157}.\\
        \midrule
        Saidulu et al. \cite{Saidulu2025} & VGG-19 & 0.1 & NS & ImageNet + & Fine-tuned for CT-specific features; \\
        & & & & Medical & $\lambda$ chosed based on hyperparameter optimization.\\
        \midrule
        Marcos et al. \cite{a18030134} & VGG-16 & 0.4 & \texttt{block1\_conv2}, \texttt{block2\_conv2} & ImageNet & $\lambda$ chosed based on ablation studies. \\
        & & & \texttt{block3\_conv3}, \texttt{block4\_conv3} & & \\
        \midrule
        Li et al. \cite{li2025ctmambahybridconvolutionalstate} & ResNet-50 & 0.01 & Last layer of its & ImageNet & The choice of $\lambda$ lacks explanation.\\
        & & & four main stages & & \\
        \midrule
        Huang et al.\cite{HUANG2021104} & VGG & 0.001 & NS & ImageNet & $\lambda$ choice is not explained. \\
        \bottomrule
        \multicolumn{6}{p{\dimexpr\textwidth-2\tabcolsep\relax}}{
        \footnotesize NS: Not Specified in the paper.
        } \\
    \end{tabular}
    }
\end{sidewaystable}

While the values for $\phi$, $\lambda$, $c$, and $\mathcal{E}$ presented in Table~\ref{tab:related_works} are widely adopted across the literature, their selection is often based on precedent or empirical observations from pioneering works in other image processing domains. To the best of our knowledge, there is a notable lack of systematic investigations or comprehensive ablation studies within the LDCT denoising context that rigorously justify these particular choices over alternatives. For instance, whether the optimal feature layers differ between medical and natural images remains to be explored, as does the possibility that the ideal weighting for the perceptual loss varies depending on the task or dataset. This raises the concern that the prevalent use of these specific parameters may, in some cases, be more arbitrary or conventional rather than strictly evidence-based from dedicated comparative studies within the LDCT application space, potentially leaving room for further optimization.

\section{Methodology}
\label{sec:methodology}

To address our research questions, we vary the encoder’s feature extraction depth $\phi$ and its pre-training dataset $c$, and calibrate the loss weight $\lambda$ to control the relative influence of the perceptual term during training. Controlling this influence is, to the best of our knowledge, a novel methodological step not previously explored in the literature, yet it plays a central role in design optimization. We argue that evaluating different configurations of perceptual loss without accounting for their relative contribution to the total loss may lead to misleading conclusions, as identical values of $\lambda$ can induce substantially different optimization dynamics depending on the feature layer and pre-training context used.

The complete experimental procedure is detailed in Algorithm~\ref{alg:methodology}, where four experimental configurations were designed using a fixed VGG-19 encoder \(\mathcal{E}\). The feature layers $\phi$ considered include \texttt{block3\_conv2}, which captures mid-level visual patterns, and \texttt{block5\_conv4}, which encodes higher-level semantic content. As for the pre-training datasets $c$, we consider both \texttt{ImageNet}, which yields general-purpose visual features, and a domain-specific \texttt{Medical} dataset designed to emphasize anatomical structures. In each experimental configuration, models were trained using a selected \(\lambda\) value that equalizes the contribution of the perceptual loss to the total cost function used during backpropagation, thereby enabling fairer comparisons and more interpretable results.

\begin{algorithm}[!ht]
\caption{Experimental Methodology}
\label{alg:methodology}
\begin{algorithmic}[1]
\REQUIRE Dataset \(T\), fixed encoder \(\mathcal{E}\), network model \(f_\theta\)
\ENSURE For each experimental configuration \(E_k\), return the vectors \(\mathrm{PSNR}_{E_k}^{\gamma}\), \(\mathrm{SSIM}_{E_k}^{\gamma}\), and \(\mathrm{NRMSE}_{E_k}^{\gamma}\), containing the respective metric values for all images in the test set \(\gamma\)

\STATE \textbf{Dataset Preparation:} Preprocess dataset \(T\) to obtain input-output pairs \((\mathbf{x}, \mathbf{y})\), and split into training \(\alpha\) (70\%), validation \(\beta\) (10\%), and testing \(\gamma\) (20\%) subsets.

\STATE \textbf{Experiment Definition:} Define four experimental configurations:
\begin{itemize}
    \item \textbf{Experiment 1 ($E_1$):} \(\mathcal{E} = \texttt{VGG-19}\), \(c = \texttt{ImageNet}\), \(\phi = \texttt{block3\_conv2}\)
    \item \textbf{Experiment 2 ($E_2$):} \(\mathcal{E} = \texttt{VGG-19}\), \(c = \texttt{ImageNet}\), \(\phi = \texttt{block5\_conv4}\)
    \item \textbf{Experiment 3 ($E_3$):} \(\mathcal{E} = \texttt{VGG-19}\), \(c = \texttt{Medical}\), \(\phi = \texttt{block3\_conv2}\)
    \item \textbf{Experiment 4 ($E_4$):} \(\mathcal{E} = \texttt{VGG-19}\), \(c = \texttt{Medical}\), \(\phi = \texttt{block5\_conv4}\)
\end{itemize}

\FORALL{experiments \(E_k \in \{E_1, E_2, E_3, E_4\}\)}
    \STATE Initialize configuration \((\phi, c)\) from \(E_k\)
    \STATE \textbf{Adjust} \(\lambda\) using Algorithm~\ref{alg:estimate_p_curve} so that \(\Psi(\lambda, \phi, c) \approx 0.95\)
    \STATE Train \(f_\theta\) on training set \(\alpha\), and validate using set \(\beta\)
    \STATE Compute image metrics on test set \(\gamma\), obtaining the vectors \(\mathrm{PSNR}_{E_k}^{\gamma}\), \(\mathrm{SSIM}_{E_k}^{\gamma}\), and \(\mathrm{NRMSE}_{E_k}^{\gamma}\), each containing one value per image in \(\gamma\)
\ENDFOR
\end{algorithmic}
\end{algorithm}

We then introduce the concept of \textit{perceptual influence} \(\Psi(\lambda, \phi, c)\), which is essential to ensure that all models are trained under a consistent contribution of the perceptual loss term relative to the total loss. It is defined as the average proportion of the perceptual loss with respect to the total cost over the training set. The formal definition of $\Psi$, which depends on the selected feature layer $\phi$, the pre-training dataset $c$, and the loss weight $\lambda$, is given by Eq.~\eqref{eq:perceptual_influence}.

\begin{equation}
    \Psi(\lambda, \phi, c) = \frac{1}{N} \sum_{i=1}^N \frac{\lambda \cdot \ell^{c,\phi}_{\text{PL}}(f_\theta(\mathbf{x}^{(i)}), \mathbf{y}^{(i)})}{\ell_\text{MSE}(f_\theta(\mathbf{x}^{(i)}), \mathbf{y}^{(i)}) + \lambda \cdot \ell^{c,\phi}_{\text{PL}}(f_\theta(\mathbf{x}^{(i)}), \mathbf{y}^{(i)})}
    \label{eq:perceptual_influence}
\end{equation}

To determine the appropriate \(\lambda\) value that achieves a target perceptual influence (\(\Psi \approx 0.95\) in this work), we employ a calibration procedure described in Algorithm~\ref{alg:estimate_p_curve}. We simulate the loss behavior on the training set using an untrained model and compute \(\Psi(\lambda, \phi, c)\) across a range of \(\lambda\) values. This curve allows us to select the \(\lambda\) that meets the desired influence level before actual training begins, ensuring that all configurations are compared under equivalent training conditions with respect to the role of the perceptual loss.

\begin{algorithm}[!ht]
\caption{Estimate \(\Psi(\lambda, \phi, c)\) curve for fixed \(\mathcal{E}\), \(\phi\) and \(c\)}
\label{alg:estimate_p_curve}
\begin{algorithmic}[1]
\REQUIRE Training set \(T = \{(\mathbf{x}^{i}, \mathbf{y}^{i})\}_{i=1}^N\); encoder \(\mathcal{E}_c^\phi\); untrained model \(f_\theta\); list of \(\lambda\) values: \(\Lambda \)
\ENSURE Function values \(\Psi(\lambda, \phi, c)\) for all \(\lambda \in \Lambda\)

\STATE Initialize: \( S_\text{MSE} \leftarrow 0 \), \( S_\text{PL} \leftarrow 0 \)
\FOR{each training pair \((\mathbf{x}^{i}, \mathbf{y}^{i}) \in T\)}
    \STATE \( \hat{\mathbf{y}}^{(i)} \leftarrow f_\theta(\mathbf{x}^{i}) \)
    \STATE \( \ell_\text{MSE}^{(i)} \leftarrow \|\hat{\mathbf{y}}^{i} - \mathbf{y}^{i}\|^2 \)
    \STATE \( \ell_\text{PL}^{i} \leftarrow \|\mathcal{E}_c^\phi(\hat{\mathbf{y}}^{i}) - \mathcal{E}_c^\phi(\mathbf{y}^{i})\|^2 \)
    \STATE \( S_\text{MSE} \leftarrow S_\text{MSE} + \ell_\text{MSE}^{i} \)
    \STATE \( S_\text{PL} \leftarrow S_\text{PL} + \ell_\text{PL}^{i} \)
\ENDFOR

\FOR{each \(\lambda \in \Lambda\)}
    \STATE Compute: \( \Psi(\lambda, \phi, c) = \frac{\lambda \cdot S_\text{PL}}{S_\text{MSE} + \lambda \cdot S_\text{PL}} \)
    \STATE Store or plot \((\lambda, \Psi(\lambda, \phi, c))\)
\ENDFOR
\end{algorithmic}
\end{algorithm}

To address research question \textbf{RQ1}, we execute Algorithm~\ref{alg:estimate_p_curve} for each of the four experimental configurations (\(E_1\) through \(E_4\)), systematically quantifying how the perceptual influence varies across different representation levels $\phi$ and pre-training contexts $c$. In response to \textbf{RQ2}, we conduct a comparative analysis of the reconstruction quality by evaluating \(\mathrm{PSNR}_{E_k}^{\gamma}\), \(\mathrm{SSIM}_{E_k}^{\gamma}\), and \(\mathrm{NRMSE}_{E_k}^{\gamma}\) for all \(k \in \{1, 2, 3, 4\}\). Finally, building on the findings from \textbf{RQ1} and \textbf{RQ2}, we derive practical guidelines for selecting perceptual loss configurations, thereby addressing \textbf{RQ3}.

\section{Experiments}
\label{sec:experiments}

All experiments in this study were set up using the TensorFlow framework\footnote{\url{https://www.tensorflow.org/}}, and executed on a computing environment equipped with an NVIDIA GeForce GTX 1650 GPU, an Intel Core i5 10th-generation CPU, and 8 GB of RAM. This hardware configuration was sufficient to support the training and evaluation procedures for all experimental setups without significant resource constraints.

To ensure transparency and reproducibility, the following subsections detail the experimental settings adopted throughout the study. Subsection~\ref{sec:dataset} presents the characteristics and preprocessing steps associated with the dataset \(T\), including the partitioning strategy used for training, validation, and testing. Subsection~\ref{sec:hyperparam} describes the hyperparameter configurations applied in each of the four experimental scenarios \(E_1\) to \(E_4\), including network training parameters and perceptual loss settings. For complete access to the implementation, the code and configuration files for all experiments are publicly available in our GitHub repository: \url{https://github.com/vngabriel/perceptual-influence}.

\subsection{Dataset}
\label{sec:dataset}

To evaluate the effect of different perceptual loss configurations on low-dose CT image enhancement, we employ the \textbf{Low-Dose CT Image and Projection Data} from the \textbf{2016 NIH-AAPM-Mayo Clinic Low-Dose CT Grand Challenge}\footnote{\url{https://www.aapm.org/grandchallenge/lowdosect/}}~\cite{https://doi.org/10.1002/mp.12345}. This publicly available dataset provides paired CT scans acquired at both low and normal radiation dose levels, facilitating direct comparisons between enhanced reconstructions and high-quality ground-truth images.

The dataset contains CT volumes from 30 patients; however, only 10 of these include paired scans with both low-dose (LDCT) and normal-dose (NDCT) acquisitions, identified as: \{L067, L097, L109, L143, L192, L286, L291, L310, L333, L506\}. The remaining 20 patients were designated for challenge validation and are excluded from this study. The dataset was partitioned into training, validation, and testing subsets, adhering to established practices \cite{Eulig2024-xs,kim2025bioattanatomicalpriordriven}. Critically, to mitigate the risk of data leakage and foster reliable model generalization, the 10 paired-patient cases were allocated such that each patient's complete data was confined to a single, distinct subset, thereby ensuring no overlap between these groups \cite{7947200,Chen:17,yang2025patientlevelanatomymeetsscanninglevel}. The distribution was as follows: seven patients \{L067, L310, L286, L333, L291, L109, L143\} for training ($\alpha$) ($70\%$); one patient \{L192\} for validation ($\beta$) ($10\%$); and two patients \{L097, L506\} for testing ($\gamma$) ($20\%$).

All CT scans were provided in DICOM format and processed using the following standard pipeline: (i) slices were sorted according to patient positioning metadata; (ii) Hounsfield unit (HU) values were normalized to the \([0, 1]\) range using the transformation using \( HU_{\text{min}} = -1024\) and \(HU_{\text{max}} = 3072\) as suggested in~\cite{7947200}; and (iii) resulting volumes were stored in NumPy format for efficient access during training and evaluation.

\begin{figure}[!h]
    \centering
    \begin{minipage}{0.485\columnwidth}
        \centering
        \includegraphics[width=\linewidth]{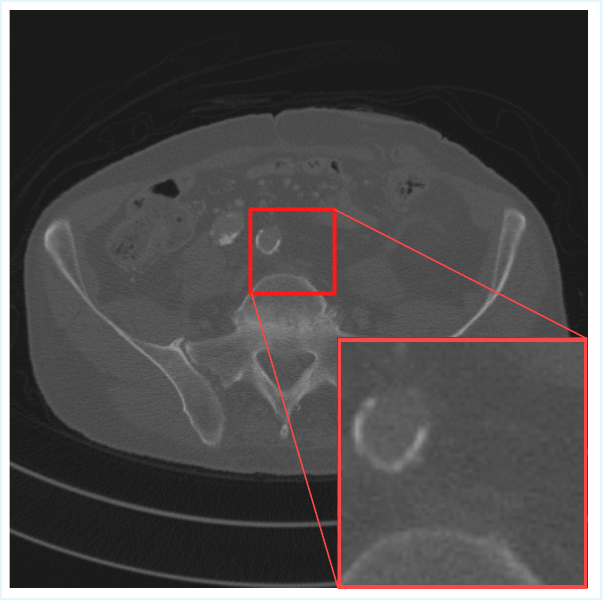}
        \caption*{(a)}
    \end{minipage}
    \hfill
    \begin{minipage}{0.48\columnwidth}
        \centering
        \includegraphics[width=\linewidth]{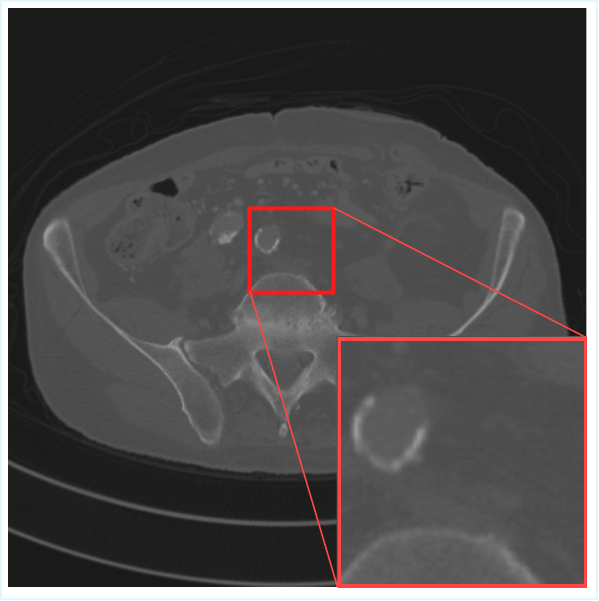}
        \caption*{(b)}
    \end{minipage}
    \caption{Images (a) and (b) show full $512 \times 512$ CT slices for LDCT and NDCT, respectively, with a highlighted $96 \times 96$ region indicated by the red box.}
    \label{fig:exemplo_mayo}
\end{figure}

This process yielded a total of 16,628 spatially aligned pairs of LDCT and NDCT slices. This pre-processing ensures consistency and comparability across experiments, allowing reliable assessment of image enhancement performance. An illustrative example of the dataset is shown in Figure~\ref{fig:exemplo_mayo}.

\subsection{Experimental Setup}
\label{sec:hyperparam}

In this work, we adopt the U-Net model \cite{ronneberger2015unetconvolutionalnetworksbiomedical} as the architecture for \(f_{\theta}\), given its widespread use and proven effectiveness in image denoising and medical image reconstruction tasks \cite{zbontar2019fastmriopendatasetbenchmarks}. To ensure the reproducibility of our experiments, we explicitly define the vector of hyperparameters \(\boldsymbol{\Theta}\) used during training. Table~\ref{tab:hp_fixos} presents the hyperparameters that remained fixed across all experiments. These values were selected based on prior empirical evidence and align with best practices reported in the literature~\cite{7947200}.

After applying Algorithm~\ref{alg:estimate_p_curve}, we determined a specific perceptual weight \(\lambda\) for each experimental configuration. Experiment~1 (\(E_1\)) was conducted with \(\lambda = 10^{-5}\), Experiment~2 (\(E_2\)) with \(\lambda = 10^{-2}\), Experiment~3 (\(E_3\)) with \(\lambda = 10^{-5}\), and Experiment~4 (\(E_4\)) with \(\lambda = 10^{-4}\). To provide a reference based on the most used $\lambda$ and $\phi$ used in the literature \cite{8964295,mahmoud2023variant,9023861,8340157}, we included additional results referred to as \textit{baseline}, which adopts the VGG-19 trained on the ImageNet with the perceptual layer $\phi$  \texttt{block5\_conv4} and $\lambda = 10^{-1}$.

\begin{table}[!ht]
    \centering
    \setlength{\tabcolsep}{2pt}
    \caption{Hyperparameter vector $\boldsymbol{\Theta}$ fixed for all experiments.}
    \label{tab:hp_fixos}
    \begin{tabular}{c p{0.6\linewidth} c}
        
        \textbf{Hyperparameter} & \textbf{Description} & \textbf{Value}\\
        \hline
            $patch\_size$ & Size of the patches extracted from the images for training. & 96 \\
            $patch\_skip$ & Step size for extracting patches while traversing the image. & 96 \\
            $batch$ & Number of samples used per iteration during training. & 16 \\
            $epochs$ & Total number of training epochs. & 10 \\
            $lr$ & Learning rate used during the CNN training. & $10^{-3}$ \\
            $\beta_1$ & First momentum parameter for the Adam optimizer. & $0.9$ \\
            $\beta_2$ & Second momentum parameter for the Adam optimizer. & $0.999$ \\
            $seed$ & Seed to ensure reproducibility in the experiments. & 5 \\
            $channels$ & Number of channels in the first U-Net layer. & 32\\
        \hline
    \end{tabular}
\end{table}


\section{Results and Discussion}
\label{sec:results}

To provide a structured analysis aligned with the study’s objectives, the results are organized into three subsections, each corresponding to one of the research questions (\textbf{RQ1}–\textbf{RQ3}). For each case, we revisit the motivation behind the question and summarize the findings derived from the experimental methodology.

\subsection{Variation of Perceptual Influence Across Perceptual Setups}
\label{sec:rq1}

\textbf{RQ 1} investigates how the perceptual influence \(\Psi(\lambda, \phi, c)\) varies as a function of the selected representation level \(\phi\) and the pre-training context \(c\), under a fixed encoder architecture \(\mathcal{E}\). By executing Algorithm~\ref{alg:estimate_p_curve} across all four experimental setups (\(E_1\) to \(E_4\)), we aimed to quantify how different semantic feature levels and training domains affect the relative contribution of the perceptual term in the total loss function. This analysis is essential for understanding the dynamics of loss balancing prior to full model training.

\begin{figure}[!h]
\centering
  \includegraphics[width=\columnwidth]{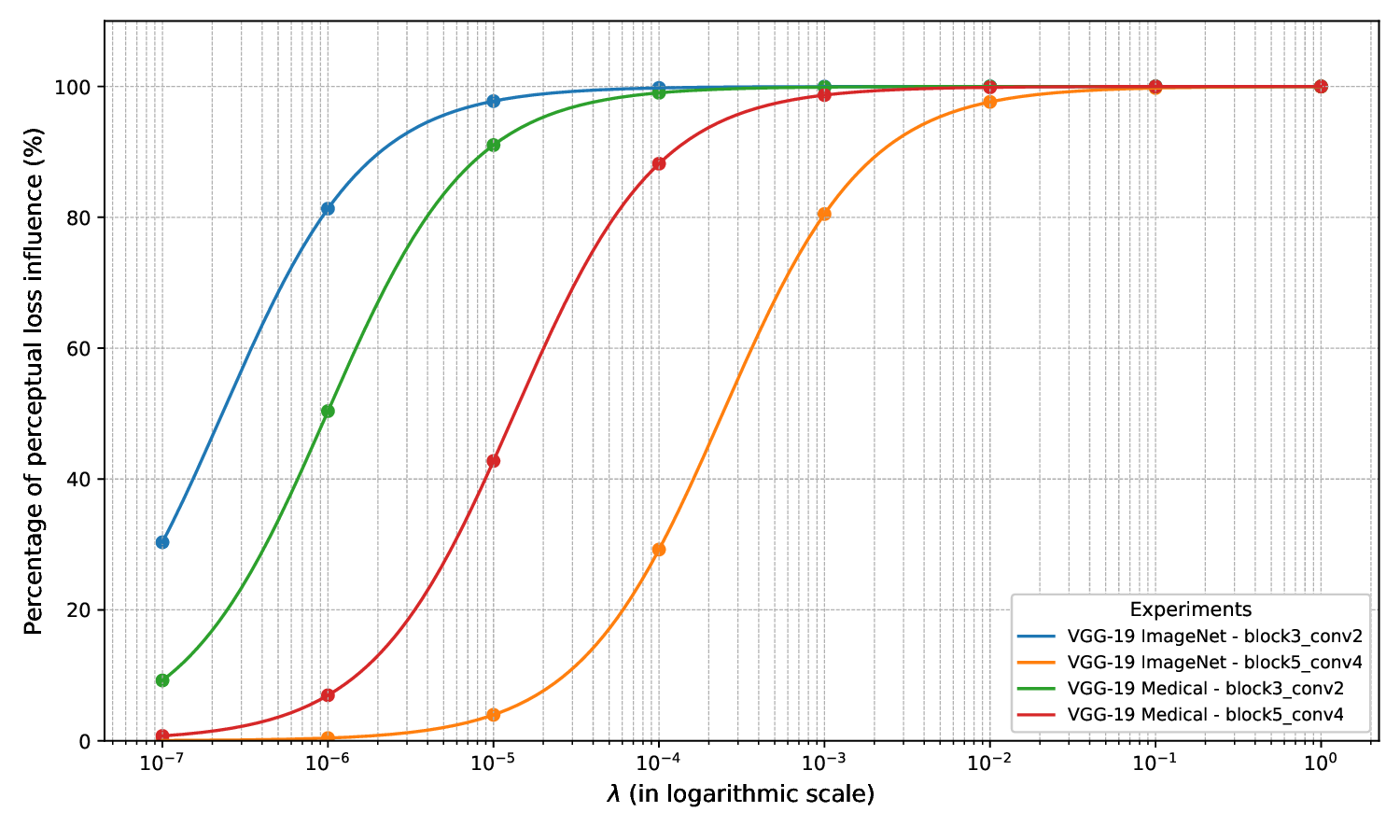}
\caption{Perceptual influence \(\Psi(\lambda, \phi, c)\) as a function of the loss weight \(\lambda\), computed for each experimental configuration \(\{E_1, E_2, E_3, E_4\}\). The curves illustrate how the contribution of the perceptual term $\ell_\text{PL}$ evolves relative to the total cost $J_\theta$ across \(\lambda\) values from $10^{-7}$ to $1$, under different combinations of feature representations \(\phi\) and pre-training contexts \(c\).}
\label{fig:perceptual_influence}

\end{figure}

From the analysis of \(\Psi(\lambda, \phi, c)\) in Figure \ref{fig:perceptual_influence}, we observe that both the feature level \(\phi\) and the pre-training context \(c\) substantially affect the balance between pixel-level fidelity and perceptual alignment. Specifically, configurations using low-level feature layers (e.g., \texttt{block3\_conv2}) tend to exhibit higher perceptual influence for the same \(\lambda\), \textbf{suggesting that low-level features contribute more strongly to the perceptual loss}. 

Furthermore, the analysis presented in Figure~\ref{fig:perceptual_influence} reveals that \textbf{the commonly adopted value \(\lambda = 0.1\) of \textit{baseline} \cite{8964295,mahmoud2023variant,9023861,8340157} is not justifiable in any of the experimental configurations evaluated in this study}. For all combinations of \(\phi\) and \(c\) considered, setting \(\lambda = 0.1\) results in a perceptual influence of approximately 100\%, meaning that the perceptual loss term \(\ell_\text{PL}\) entirely dominates the total cost function \(J_\theta\). In such a scenario, the pixel-wise component \(\ell_\text{MSE}\) in Eq.~\eqref{eq:total_loss} contributes negligibly to the optimization process, offering no practical justification for its inclusion. This finding highlights the importance of calibrating \(\lambda\) based on a principled analysis rather than relying on conventional defaults.

\subsection{Performance Comparison Across Perceptual Setups}
\label{sec:rq2}

\textbf{RQ 2} focused on comparing the LDCT enhacement performance obtained in each experimental setup, with the goal of identifying the most effective combinations of \(\phi\), \(c\), and \(\lambda\) under the constraint \(\Psi(\lambda, \phi, c) \approx 0.95\). The evaluation is based on quantitative metrics computed on the test set \(\gamma\), specifically \(\mathrm{PSNR}_{E_k}^{\gamma}\), \(\mathrm{SSIM}_{E_k}^{\gamma}\), and \(\mathrm{NRMSE}_{E_k}^{\gamma}\) for each experiment \(E_k\). This comparison provides insight into how different perceptual configurations influence the final quality of LDCT image reconstructions.

Our quantitative results are presented in Table~\ref{tab:resultado}. The best performance was achieved in Experiment~\(E_1\), which used the VGG-19 encoder $\mathcal{E}$ pre-trained on the ImageNet dataset (\(c = \texttt{ImageNet}\)) and the intermediate-level feature representation layer \(\phi = \texttt{block3\_conv2}\). It is worth noting that state-of-the-art approaches rely on deeper representations, particularly \(\phi = \texttt{block5\_conv4}\) \cite{8340157}. However, our findings suggest that \textbf{when using a perceptual loss based on a generic pre-training context such as ImageNet, guiding the training with lower-level features may lead to superior reconstruction performance}.

\begin{table}[ht]
    \centering
    \setlength{\tabcolsep}{2pt}
    \caption{Evaluation results for each experiment. Mean values with standard deviations in parentheses.}
    \label{tab:resultado}
    \begin{tabular}{c c c c c}
        \textbf{Experiment} & \textbf{SSIM} & \textbf{PSNR} & \textbf{NRMSE} & \textbf{Rank} \\
        \hline
        \textit{baseline} & 0.929 (0.053) & 38.98 (2.45) & 0.068 (0.019) & 4 \\
        \(E_1\) & \textbf{0.944 (0.043)} & 42.61 (3.87) & 0.048 (0.021) & 1 \\
        \(E_2\) & 0.933 (0.050) & 41.52 (3.62) & 0.054 (0.022) & 3 \\
        \(E_3\) & 0.899 (0.062) & 39.70 (3.40) & 0.065 (0.026) & 5 \\
        \(E_4\) & 0.937 (0.045) & 41.80 (3.39) & 0.051 (0.020) & 2 \\
        \hline
    \end{tabular}
\end{table}

\begin{figure}[!h]
\centering
  \includegraphics[width=\textwidth]{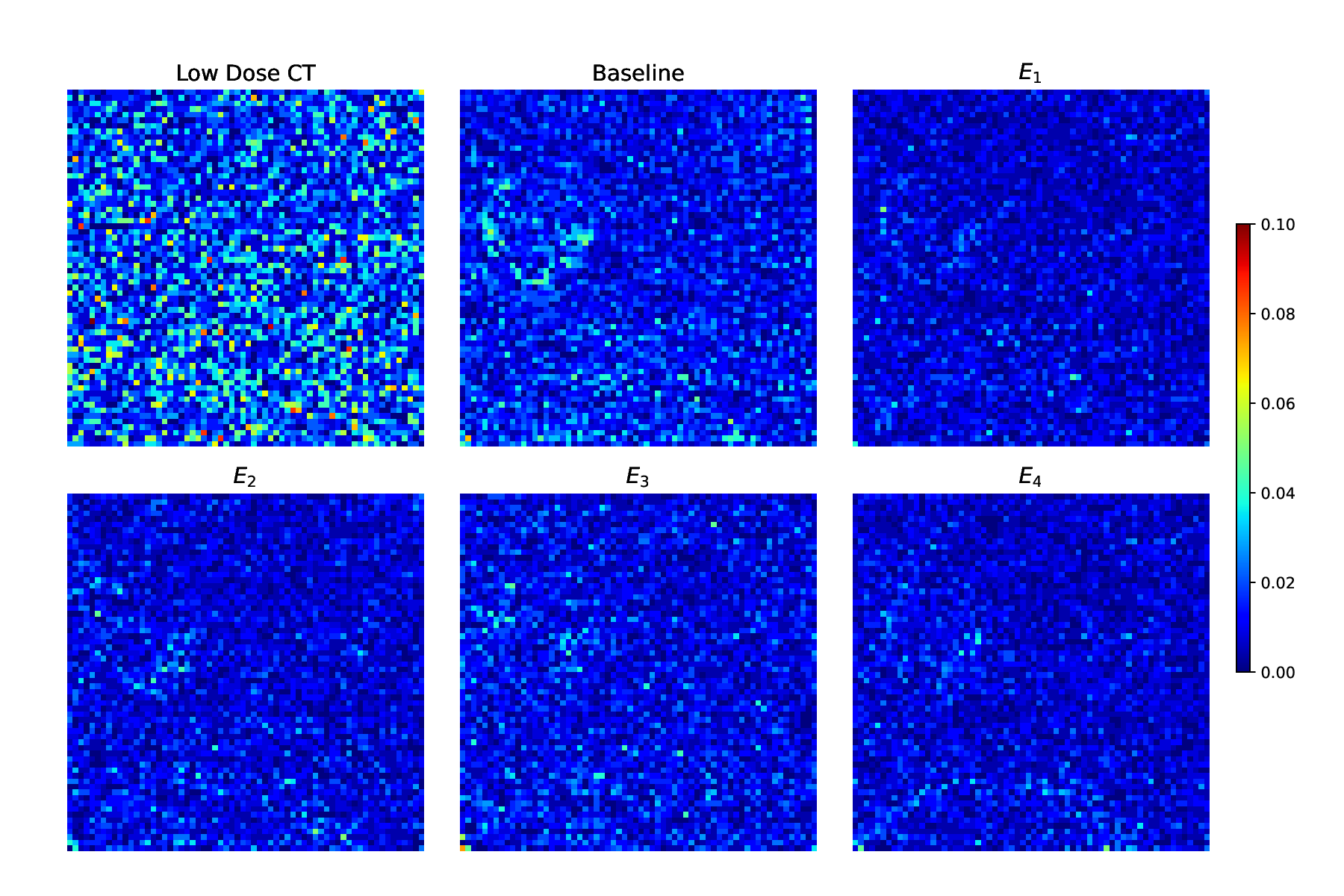}
  \caption{Error heatmaps between denoised outputs and ground truth for each experimental configuration.}
  \label{fig:heatmap2}
\end{figure}

From a qualitative perspective, the darker heatmaps corresponding to Experiments~E\(_1\) and E\(_4\), shown in Figure~\ref{fig:heatmap2}, visually support the ranking indicated by the quantitative metrics. In Experiment \(E_4\), the configuration involved a VGG-19 pre-trained on a medical dataset (\(c = \texttt{Medical}\)) and the deep representation level \(\phi = \texttt{block5\_conv4}\). This suggests that \textbf{when domain-specific pre-training is available, leveraging higher-level semantic features can be beneficial in guiding network optimization}.

Although our results are reported across three metrics - NRMSE, PSNR, and SSIM - it is important to note that both NRMSE\footnote{NRMSE is defined as \(\text{NRMSE} = \frac{\| \hat{\mathbf{y}} - \mathbf{y} \|_2}{\| \mathbf{y} \|_2}\)} and PSNR\footnote{PSNR is defined as \(\text{PSNR} = 20 \cdot \log_{10} \left( \frac{\text{MAX}_\mathbf{y}}{\sqrt{\text{MSE}}} \right)\)} are pixel-wise similarity metrics. Therefore, in the rest of this section, we focus on the SSIM distribution across the test set. Table~\ref{tab:teste_hipotese_p1} presents the results of the statistical analysis used to compare SSIM values across experimental configurations. We employed the Wilcoxon signed-rank test for paired samples, using a significance level of \(\alpha = 0.05\). The null hypothesis (\(H_0\)) assumes that the median difference in SSIM between the paired experimental conditions is zero, indicating no effect. The alternative hypothesis (\(H_1\)) assumes that the median difference is not zero, indicating a statistically significant difference between the experimental configurations.

\begin{table}[!ht]
    \centering
    \caption{Hypothesis testing between Experiment 1 (best result) and others. Statistical significance (p-value) and effect size (Cohen’s \(d\)).}
    \label{tab:teste_hipotese_p1}
    \begin{tabular}{c c c c c}
  
        \textbf{Comparison} & \textbf{SSIMs} & \textbf{p-value} & \textbf{Effect size} & \textbf{Significant} \\
        \hline
        $E_1$ vs \textit{baseline} & 0.9443 vs 0.9293 & 0.0 & 0.31 & Yes \\
        $E_1$ vs $E_2$ & 0.9443 vs 0.9333 & 0.0 & 0.24 & Yes \\
        $E_1$ vs $E_3$ & 0.9443 vs 0.8987 & 0.0 & 0.86 & Yes \\
        $E_1$ vs $E_4$ & 0.9443 vs 0.9370 & 0.0 & 0.17 & Yes \\
        \hline
    \end{tabular}
\end{table}

The statistical tests reported in Table~\ref{tab:teste_hipotese_p1} reveal significant differences in SSIM across all experimental configurations, reinforcing the importance of carefully setting the perceptual loss parameters. These findings emphasize that the choice of feature level \(\phi\), pre-training context \(c\), and weighting factor \(\lambda\) plays a critical role in reconstruction quality. Therefore, the concept of perceptual influence emerges as a valuable tool for guiding such configurations. It provides a principled way to compare \((\phi, c, \lambda)\) tuples that share similar influence levels.

\subsection{Guidelines for Perceptual Loss Configuration}
\label{sec:rq3}

\textbf{RQ 3} aims to derive objective, empirically grounded guidelines for selecting suitable perceptual loss configurations, based on the patterns observed in \textbf{RQ1} and \textbf{RQ2}. The goal is to assist future researchers and practitioners in choosing appropriate values for \(\phi\), \(c\), and \(\lambda\) that maximize performance while maintaining a desirable level of perceptual influence. This subsection synthesizes the previous findings to propose actionable recommendations supported by numerical evidence.
 
The results demonstrate a clear interaction between the pre-training context \(c\) and the optimal feature representation level \(\phi\). When using a VGG-19 pre-trained on ImageNet, the initial layers (\texttt{block3\_conv2}) led to better reconstruction performance, likely due to their emphasis on low-level features such as edges and textures, which generalize well across domains. In contrast, when the encoder was fine-tuned on medical imaging data, deeper layers (\texttt{block5\_conv4}) produced superior results, suggesting that high-level semantic representations aligned more closely with domain-specific anatomical structures. These findings indicate that the selection of \(\phi\) and \(\lambda\) should not follow a one-size-fits-all strategy. In practical terms, our results suggest the following rule of thumb: \textbf{when using encoders trained on generic datasets such as ImageNet, perceptual losses should prioritize lower-level feature representations. Conversely, when encoders are trained or fine-tuned on domain-specific data, higher-level features become more informative and should be preferred in guiding the enhancement of LDCT images}.

\section{Conclusion}
\label{sec:conclusion}

This study examined how different perceptual loss configurations affect LDCT image enhancement and introduced the \emph{perceptual influence} metric to quantify the relative contribution of the perceptual term $\ell_\text{PL}$ within the total loss $J_\theta$. Our results demonstrate that the choices of feature level \(\phi\) and pre-training context \(c\) have a substantial impact on image quality. Furthermore, we found that improper selection of the weighting parameter \(\lambda\) - such as the commonly adopted but empirically unjustified value \(\lambda = 0.1\) - can lead to the dominance of $\ell_\text{PL}$, effectively nullifying the influence of the pixel-wise loss $\ell_{\text{MSE}}$. Additionally, using the proposed framework, we observed that when encoders are pre-trained on natural datasets (e.g., ImageNet), lower-level features tend to perform better, whereas domain-specific pre-training favors higher-level semantic representations.


\section*{Acknowledgments}

This project was partially funded by the Brazilian agency CNPq (Conselho Nacional de Desenvolvimento Científico e Tecnológico) (grant number 403477/2024-7).

This research also received funding from the Flemish Government under the “Onderzoeksprogramma Artificiële Intelligentie (AI) Vlaanderen” program and the Fund for Scientific Research Flanders (FWO under grant number G096324N).

\section*{Conflicts of interest}
The authors declare that they have no conflicts to disclose.

\section*{Ethics Statement}

This study did not involve any new data collection from human participants and was therefore not subject to institutional ethical review. The research was conducted using the publicly available \textbf{2016 NIH-AAPM-Mayo Clinic Low-Dose CT Grand Challenge}\footnote{\url{https://www.aapm.org/grandchallenge/lowdosect/}}~\cite{https://doi.org/10.1002/mp.12345} dataset, hosted by the American Association of Physicists in Medicine (AAPM). The dataset consists of retrospectively collected abdominal CT scans acquired during clinically indicated exams. The data collection was approved by the authors' Institutional Review Board (IRB), performed in compliance with HIPAA regulations, and conducted with a waiver of informed consent. All data were fully anonymized before being shared, and no identifiable patient information is present in the dataset.

\section*{Declaration of Generative AI and AI-assisted Technologies in the Writing Process}

During the preparation of this work, the author(s) used ChatGPT (GPT-4o) and Google Gemini (Gemini 2.5 Pro) to improve the readability of the English text. After using these tools, the author(s) thoroughly reviewed and edited the content as needed and take(s) full responsibility for the content of the publication.

\bibliographystyle{elsarticle-num.bst}
\bibliography{references}

\begin{thebibliography}{10}
\expandafter\ifx\csname url\endcsname\relax
  \def\url#1{\texttt{#1}}\fi
\expandafter\ifx\csname urlprefix\endcsname\relax\def\urlprefix{URL }\fi
\expandafter\ifx\csname href\endcsname\relax
  \def\href#1#2{#2} \def\path#1{#1}\fi

\bibitem{7947200}
H.~Chen, Y.~Zhang, M.~K. Kalra, F.~Lin, Y.~Chen, P.~Liao, J.~Zhou, G.~Wang, Low-dose {CT} with a residual encoder-decoder convolutional neural network, IEEE Transactions on Medical Imaging 36~(12) (2017) 2524--2535.
\newblock \href {https://doi.org/10.1109/TMI.2017.2715284} {\path{doi:10.1109/TMI.2017.2715284}}.

\bibitem{8964295}
M.~Li, W.~Hsu, X.~Xie, J.~Cong, W.~Gao, {SACNN}: Self-attention convolutional neural network for low-dose {CT} denoising with self-supervised perceptual loss network, IEEE Transactions on Medical Imaging 39~(7) (2020) 2289--2301.
\newblock \href {https://doi.org/10.1109/TMI.2020.2968472} {\path{doi:10.1109/TMI.2020.2968472}}.

\bibitem{ZHANG2024108112}
J.~Zhang, W.~Gong, L.~Ye, F.~Wang, Z.~Shangguan, Y.~Cheng, \href{https://www.sciencedirect.com/science/article/pii/S0010482524001963}{A review of deep learning methods for denoising of medical low-dose {CT} images}, Computers in Biology and Medicine 171 (2024) 108112.
\newblock \href {https://doi.org/https://doi.org/10.1016/j.compbiomed.2024.108112} {\path{doi:https://doi.org/10.1016/j.compbiomed.2024.108112}}.
\newline\urlprefix\url{https://www.sciencedirect.com/science/article/pii/S0010482524001963}

\bibitem{8340157}
Q.~Yang, P.~Yan, Y.~Zhang, H.~Yu, Y.~Shi, X.~Mou, M.~K. Kalra, Y.~Zhang, L.~Sun, G.~Wang, Low-dose {CT} image denoising using a generative adversarial network with wasserstein distance and perceptual loss, IEEE Transactions on Medical Imaging 37~(6) (2018) 1348--1357.
\newblock \href {https://doi.org/10.1109/TMI.2018.2827462} {\path{doi:10.1109/TMI.2018.2827462}}.

\bibitem{johnson2016perceptuallossesrealtimestyle}
J.~Johnson, A.~Alahi, L.~Fei-Fei, \href{https://arxiv.org/abs/1603.08155}{Perceptual losses for real-time style transfer and super-resolution} (2016).
\newblock \href {http://arxiv.org/abs/1603.08155} {\path{arXiv:1603.08155}}.
\newline\urlprefix\url{https://arxiv.org/abs/1603.08155}

\bibitem{8099502}
C.~Ledig, L.~Theis, F.~Huszár, J.~Caballero, A.~Cunningham, A.~Acosta, A.~Aitken, A.~Tejani, J.~Totz, Z.~Wang, W.~Shi, Photo-realistic single image super-resolution using a generative adversarial network, in: 2017 IEEE Conference on Computer Vision and Pattern Recognition (CVPR), 2017, pp. 105--114.
\newblock \href {https://doi.org/10.1109/CVPR.2017.19} {\path{doi:10.1109/CVPR.2017.19}}.

\bibitem{li2025ctmambahybridconvolutionalstate}
L.~Li, W.~Wei, L.~Yang, W.~Zhang, J.~Dong, Y.~Liu, H.~Huang, W.~Zhao, \href{https://arxiv.org/abs/2411.07930}{{CT}-mamba: A hybrid convolutional state space model for low-dose {CT} denoising} (2025).
\newblock \href {http://arxiv.org/abs/2411.07930} {\path{arXiv:2411.07930}}.
\newline\urlprefix\url{https://arxiv.org/abs/2411.07930}

\bibitem{mahmoud2023variant}
A.~A. Mahmoud, H.~A. Sayed, S.~S. Mohamed, Variant wasserstein generative adversarial network applied on low dose {CT} image denoising., Computers, Materials \& Continua 75~(2) (2023).

\bibitem{10535288}
N.~Saidulu, P.~R. Muduli, A.~Dasgupta, Rhlnet: Robust hybrid loss-based network for low-dose {CT} image denoising, IEEE Transactions on Instrumentation and Measurement 73 (2024) 1--11.
\newblock \href {https://doi.org/10.1109/TIM.2024.3403187} {\path{doi:10.1109/TIM.2024.3403187}}.

\bibitem{9023861}
N.~T. Trung, T.~Dinh~Hoan, N.~L. Trung, L.~Manh~Ha, Robust denoising of low-dose {CT} images using convolutional neural networks, in: 2019 6th NAFOSTED Conference on Information and Computer Science (NICS), 2019, pp. 506--511.
\newblock \href {https://doi.org/10.1109/NICS48868.2019.9023861} {\path{doi:10.1109/NICS48868.2019.9023861}}.

\bibitem{electronics13163184}
W.~Zhang, A.~Salmi, C.~Yang, F.~Jiang, \href{https://www.mdpi.com/2079-9292/13/16/3184}{Innovative noise extraction and denoising in low-dose {CT} using a supervised deep learning framework}, Electronics 13~(16) (2024).
\newblock \href {https://doi.org/10.3390/electronics13163184} {\path{doi:10.3390/electronics13163184}}.
\newline\urlprefix\url{https://www.mdpi.com/2079-9292/13/16/3184}

\bibitem{HUANG2021104}
Z.~Huang, X.~Liu, R.~Wang, J.~Chen, P.~Lu, Q.~Zhang, C.~Jiang, Y.~Yang, X.~Liu, H.~Zheng, D.~Liang, Z.~Hu, \href{https://www.sciencedirect.com/science/article/pii/S0925231220316672}{Considering anatomical prior information for low-dose {CT} image enhancement using attribute-augmented wasserstein generative adversarial networks}, Neurocomputing 428 (2021) 104--115.
\newblock \href {https://doi.org/https://doi.org/10.1016/j.neucom.2020.10.077} {\path{doi:https://doi.org/10.1016/j.neucom.2020.10.077}}.
\newline\urlprefix\url{https://www.sciencedirect.com/science/article/pii/S0925231220316672}

\bibitem{you2018structurally}
C.~You, Q.~Yang, H.~Shan, L.~Gjesteby, G.~Li, S.~Ju, Z.~Zhang, Z.~Zhao, Y.~Zhang, W.~Cong, G.~Wang, Structurally-sensitive multi-scale deep neural network for low-dose {CT} denoising, IEEE Access 6 (2018) 41839--41855.
\newblock \href {https://doi.org/10.1109/ACCESS.2018.2858196} {\path{doi:10.1109/ACCESS.2018.2858196}}.

\bibitem{simonyan2015deepconvolutionalnetworkslargescale}
K.~Simonyan, A.~Zisserman, \href{https://arxiv.org/abs/1409.1556}{Very deep convolutional networks for large-scale image recognition} (2015).
\newblock \href {http://arxiv.org/abs/1409.1556} {\path{arXiv:1409.1556}}.
\newline\urlprefix\url{https://arxiv.org/abs/1409.1556}

\bibitem{5206848}
J.~Deng, W.~Dong, R.~Socher, L.-J. Li, K.~Li, L.~Fei-Fei, Imagenet: A large-scale hierarchical image database, in: 2009 IEEE Conference on Computer Vision and Pattern Recognition, 2009, pp. 248--255.
\newblock \href {https://doi.org/10.1109/CVPR.2009.5206848} {\path{doi:10.1109/CVPR.2009.5206848}}.

\bibitem{10.3389/frsip.2021.812193}
L.~Marcos, J.~Alirezaie, P.~Babyn, \href{https://www.frontiersin.org/journals/signal-processing/articles/10.3389/frsip.2021.812193}{Low dose {CT} denoising by resnet with fused attention modules and integrated loss functions}, Frontiers in Signal Processing Volume 1 - 2021 (2022).
\newblock \href {https://doi.org/10.3389/frsip.2021.812193} {\path{doi:10.3389/frsip.2021.812193}}.
\newline\urlprefix\url{https://www.frontiersin.org/journals/signal-processing/articles/10.3389/frsip.2021.812193}

\bibitem{Saidulu2025}
N.~Saidulu, P.~R. Muduli, \href{https://doi.org/10.1007/s00034-025-03079-9}{Dynamic perception-oriented low-dose {CT} image denoising network using structure-aware self-similarity}, Circuits, Systems, and Signal Processing (Apr 2025).
\newblock \href {https://doi.org/10.1007/s00034-025-03079-9} {\path{doi:10.1007/s00034-025-03079-9}}.
\newline\urlprefix\url{https://doi.org/10.1007/s00034-025-03079-9}

\bibitem{a18030134}
L.~Marcos, P.~Babyn, J.~Alirezaie, \href{https://www.mdpi.com/1999-4893/18/3/134}{Edge detection attention module in pure vision transformer for low-dose x-ray computed tomography image denoising}, Algorithms 18~(3) (2025).
\newblock \href {https://doi.org/10.3390/a18030134} {\path{doi:10.3390/a18030134}}.
\newline\urlprefix\url{https://www.mdpi.com/1999-4893/18/3/134}

\bibitem{https://doi.org/10.1002/mp.12345}
C.~H. McCollough, A.~C. Bartley, R.~E. Carter, B.~Chen, T.~A. Drees, P.~Edwards, D.~R. Holmes~III, A.~E. Huang, F.~Khan, S.~Leng, K.~L. McMillan, G.~J. Michalak, K.~M. Nunez, L.~Yu, J.~G. Fletcher, \href{https://aapm.onlinelibrary.wiley.com/doi/abs/10.1002/mp.12345}{Low-dose {CT} for the detection and classification of metastatic liver lesions: Results of the 2016 low dose {CT} grand challenge}, Medical Physics 44~(10) (2017) e339--e352.
\newblock \href {https://doi.org/https://doi.org/10.1002/mp.12345} {\path{doi:https://doi.org/10.1002/mp.12345}}.
\newline\urlprefix\url{https://aapm.onlinelibrary.wiley.com/doi/abs/10.1002/mp.12345}

\bibitem{Eulig2024-xs}
E.~Eulig, B.~Ommer, M.~Kachelrie{\ss}, Benchmarking deep learning-based low-dose {CT} image denoising algorithms, Med. Phys. 51~(12) (2024) 8776--8788.

\bibitem{kim2025bioattanatomicalpriordriven}
N.~Kim, U.~Cho, \href{https://arxiv.org/abs/2504.01662}{Bioatt: Anatomical prior driven low-dose {CT} denoising} (2025).
\newblock \href {http://arxiv.org/abs/2504.01662} {\path{arXiv:2504.01662}}.
\newline\urlprefix\url{https://arxiv.org/abs/2504.01662}

\bibitem{Chen:17}
H.~Chen, Y.~Zhang, W.~Zhang, P.~Liao, K.~Li, J.~Zhou, G.~Wang, \href{https://opg.optica.org/boe/abstract.cfm?URI=boe-8-2-679}{Low-dose {CT} via convolutional neural network}, Biomed. Opt. Express 8~(2) (2017) 679--694.
\newblock \href {https://doi.org/10.1364/BOE.8.000679} {\path{doi:10.1364/BOE.8.000679}}.
\newline\urlprefix\url{https://opg.optica.org/boe/abstract.cfm?URI=boe-8-2-679}

\bibitem{yang2025patientlevelanatomymeetsscanninglevel}
Z.~Yang, Y.~Chen, Z.~Wang, H.~Shan, Y.~Chen, Y.~Zhang, \href{https://arxiv.org/abs/2503.00908}{Patient-level anatomy meets scanning-level physics: Personalized federated low-dose {CT} denoising empowered by large language model} (2025).
\newblock \href {http://arxiv.org/abs/2503.00908} {\path{arXiv:2503.00908}}.
\newline\urlprefix\url{https://arxiv.org/abs/2503.00908}

\bibitem{ronneberger2015unetconvolutionalnetworksbiomedical}
O.~Ronneberger, P.~Fischer, T.~Brox, \href{https://arxiv.org/abs/1505.04597}{U-net: Convolutional networks for biomedical image segmentation} (2015).
\newblock \href {http://arxiv.org/abs/1505.04597} {\path{arXiv:1505.04597}}.
\newline\urlprefix\url{https://arxiv.org/abs/1505.04597}

\bibitem{zbontar2019fastmriopendatasetbenchmarks}
J.~Zbontar, F.~Knoll, A.~Sriram, T.~Murrell, Z.~Huang, M.~J. Muckley, A.~Defazio, R.~Stern, P.~Johnson, M.~Bruno, M.~Parente, K.~J. Geras, J.~Katsnelson, H.~Chandarana, Z.~Zhang, M.~Drozdzal, A.~Romero, M.~Rabbat, P.~Vincent, N.~Yakubova, J.~Pinkerton, D.~Wang, E.~Owens, C.~L. Zitnick, M.~P. Recht, D.~K. Sodickson, Y.~W. Lui, \href{https://arxiv.org/abs/1811.08839}{fast{MRI}: An open dataset and benchmarks for accelerated mri} (2019).
\newblock \href {http://arxiv.org/abs/1811.08839} {\path{arXiv:1811.08839}}.
\newline\urlprefix\url{https://arxiv.org/abs/1811.08839}

\end{thebibliography}

\end{document}